# UPDATING THE SILENT SPEECH CHALLENGE BENCHMARK WITH DEEP LEARNING

*Yan Ji, Licheng Liu, Hongcui Wang, Zhilei Liu, Zhibin Niu, Bruce Denby*

Tianjin University, Tianjin, China

**ABSTRACT**

The 2010 Silent Speech Challenge benchmark is updated with new results obtained in a Deep Learning strategy, using the same input features and decoding strategy as in the original article. A Word Error Rate of 6.4% is obtained, compared to the published value of 17.4%. Additional results comparing new auto-encoder-based features with the original features at reduced dimensionality, as well as decoding scenarios on two different language models, are also presented. The Silent Speech Challenge archive has been updated to contain both the original and the new auto-encoder features, in addition to the original raw data.

## 1. INTRODUCTION

### 1.1. Silent speech interfaces and challenges

A Silent Speech Interface, or SSI, is defined as a device enabling speech processing in the absence of an exploitable audio signal – for example, speech recognition obtained exclusively from video images of the mouth, or from electromyographic sensors (EMA) glued to the tongue. Classic applications targeted by SSIs include:

1) Voice-replacement for persons who have lost the ability to vocalize through illness or an accident, yet who retain the ability to articulate;
2) Speech communication in environments where silence is either necessary or desired: responding to cellphone in meetings or public places without disturbing others; avoiding interference in call centers, conferences and classrooms; private communications by police, military, or business personnel.

The SSI concept was first identified as an outgrowth of speech production research, in tandem with the proliferation of the use of cellular telephones, in 2010 in a special issue of Speech Communication [1], where SSIs based on seven different non-acoustic sensor types were presented:

1)	MHz range medical ultrasound (US) + video imaging of tongue and lips;
2)	Surface electromyography, sEMG, sensors applied to the face and neck;
3)	Electromagnetic articulography EMA sensors attached to tongue, lips, jaw;
4)	Vibration sensors placed on the head and neck;
5)	Non-audible murmur microphones, NAM, placed on the neck;
6)	Electro-encephalography, EEG, electrodes;
7)	Cortical implants for a "thought-driven" SSI.

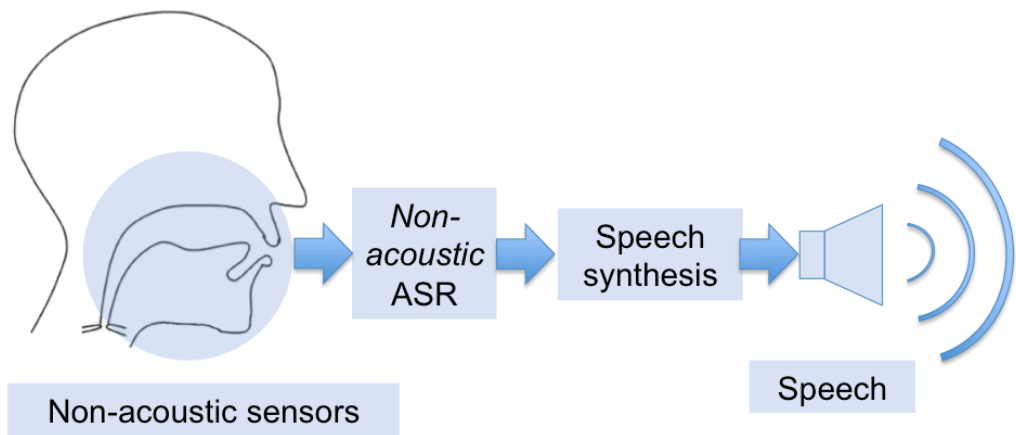

Figure 1: Overview of an SSI, showing non-acoustic sensors and non-acoustic automatic speech recognition, ASR, which can be followed by speech synthesis, or retained as a phonetic, text, or other digital representation, depending on the application.

As a non-acoustic technology, SSIs initially stood somewhat apart from the main body of speech processing, where the standard techniques are intrinsically associated with an audio signal. Nevertheless, the novelty of the SSI concept and their exciting range of applications – perhaps aided by an accrued interest in multi-modal speech processing – are gradually allowing SSI technology to join the speech processing main stream. Activity in SSI research has remained strong since the publication of [1], which received the ISCA/Eurasip Best Paper Award in 2015. A recent survey of the literature reveals dozens of publications on SSI systems, using not only on the original seven non-acoustic technologies mentioned above, but also two additional ones, namely, low frequency air-borne ultrasound; and micropower radar [1-51].

Despite this activity, SSIs today remain for the most part specialized laboratory instruments. The performance of any automatic speech recognition (ASR) system is most often characterized by a Word Error Rate, or WER, expressed as a percentage of the total number of words appearing in a corpus. To date, no SSI ASR system has been able to achieve WER parity with state-of-the-art acoustic ASR. Indeed, a number of practical issues make SSI ASR systems considerably more involved to implement than their acoustic counterparts:

1. **Sensor handling**. While in acoustic ASR this may amount to no more than routine microphone protocol, SSIs' non-acoustic sensors are often rather specialized (and expensive), and require physical contact with, or at a minimum careful placement with respect to, the speech biosignal-producing organs. This introduces problems of invasiveness; non-portability; and non-repeatability of sensor placement, bringing added complexity to SSI experiments.
2. **Interference**. An SSI should in principle be silent, but certain SSI modalities – vibration sensors, radar, and low frequency air-borne ultrasound, for example – are actually associated with signals that can propagate beyond the area of utilization of the SSI. The possibility of interference or interception may limit the adoption of these modalities outside the laboratory.
3. **Feature extraction**. While easily calculated Mel Frequency Cepstral Coefficients, MFCC, have been the acoustic ASR features of choice for decades, feature selection for the specialized sensors of SSIs remains an open question, particularly since many SSI modalities – ultrasound imaging, or EEG, for example – are of much higher intrinsic dimensionality than a simple acoustic signal. Furthermore, while the identification of stable phonetic signatures in acoustic data is today a mature field, the existence of salient landmarks in speech biosignals – arising from imaging modalities or electromyography, for example – is less evident.

Medical US operating in the MHz frequency range does not propagate outside the body. It is a well established [53] and documented [54] technique in speech production and speech pathology research, whose first use in the context of SSIs was discussed in [55]. US is a also relatively non-invasive modality, requiring only a transducer placed under the speaker's chin, coupled with a small video camera in front of the mouth to capture lip movement. These sensors can be easily accommodated in a lightweight acquisition helmet, thus minimizing sensor placement issues. US tongue imaging, with added lip video, is thus in many ways an attractive modality for building a practical SSI.

1.2. **The Silent Speech Challenge benchmark**

In 2010, an US + lip video SSI trained on the well-known TIMIT corpus achieved, with the aid of a language model (LM), a single speaker WER of 17.4% (84.2% "correct" word rate) on an independent test corpus [52], representing a promising early SSI result on a benchmark continuous speech recognition task. Subsequently, the raw image data of [52], that is, the original tongue ultrasound and lip videos, were made available online as the so-called Silent Speech Challenge, or SSC archive [56]. The purpose of the archive is to provide a stable data set to which newly developed feature extraction and speech recognition techniques can be applied. The SSC data will serve as the basis of all the experiments reported in this article.

Although a 17.4% WER for an SSI trained on a mono-speaker TIMIT corpus is "promising", it must be remembered that standard acoustic ASR can obtain similar or superior scores after training on the full **multi-speaker** acoustic TIMIT corpus, a much more challenging task. Further advances are thus still necessary in order to truly put Silent Speech Recognition, SSR, on a par with acoustic ASR.

In the past several years, improvements in acoustic speech recognition using Deep Neural Network-Hidden Markov Model (DNN-HMM) systems, rather than the traditional Gaussian Mixture Model-HMM (GMM-HMM), have become common. In this approach, a deep learning strategy is used to improve estimation of the emission probabilities of the HMM used for speech decoding. It is natural to ask to what extent a DNN-HMM approach can improve SSR performance as well. Despite the SSI implementation challenges outlined earlier, applications of deep learning techniques to SSR have indeed begun to appear. In [57], for example, tests are reported of phonetic feature discrimination for an EMG-based SSI, without a LM, on a small, experiment-specific speech corpus. In [58], deep learning on an EMA based SSI is explored, giving SSR *phone* error rates, PER, (which will be *lower* than WER) of

36%, when the Mocha-TIMIT corpus is used for training, testing, and the development of a specific bigram LM. In [59], a DNN-HMM is applied to the SSC benchmark data, albeit with a 38% WER, in a study comparing the efficacy of different feature extraction methods.

The present article reports the first application of the DNN-HMM approach to the SSC recognition benchmark using the same input features and decoding strategy as those reported in [52], thus allowing a direct comparison of performances. The SSR results obtained here are significantly improved compared to the archive, giving, in the best scenario, a 6.4% WER (94.1% "correct" word recognition rate), or a nearly threefold improvement over the benchmark value. In contrast to [57-58], furthermore, the LM used in [52], also employed here, was developed on a completely independent speech corpus. In adition, results with a second, less task-specific LM are included in the present article. Finally, tests of reduced dimensionality feature vectors, as well as completely new input features created from raw SSC archive data, are also reported here. All new features have been added to the SSC archive for future use by other researchers.

In the remainder of the article, the details of the SSC data acquisition system and a description of the available archive data are first summarized, in Section 2. Section 3 then describes the feature extraction strategy developed for the present study; while full details of the DNN-HMM based speech recognition procedure appear in Section 4. The results are summarized in Section 5, and some conclusions and perspectives for future work outlined in the final section.

## 2. SSC DATA ACQUISITION AND ARCHIVE RESOURCES

The SSC data acquisition system consisted of an acquisition helmet holding a 128 element, 4-8 MHz US probe for tongue imaging, and a black and white, infrared-illuminated video camera to capture the lips. The 320×240 pixel tongue images and 640×480 pixel lip images created by the system were acquired in a synchronized manner at 60 frames per second (fps) using the Ultraspeech multisensory acquisition system [60].

The SSC training corpus consists of US and lip video data from a single native English speaker pronouncing the 2342 utterances (47 lists of 50 sentences) of the TIMIT corpus, in the non-verbalized punctuation manner. The speech was recorded silently, i.e., without any vocalization; there is therefore no audio track. The test set is comprised of one hundred short sentences selected from the WSJ0 5000-word corpus [61] read by the same speaker. The data are available at the web address indicated in [56]. The archive initially contained only the raw ultrasound and lip images of the training and test sets; the original features used, as well as the reduced-length feature vectors and new features created for the present article (see section 3), have now been appended to it. Speech recognition for the Challenge data was carried out in a standard GMM-HMM scheme and made use of a LM, which is also included in the archive. Further details appear in section 4.

## 3. FEATURE EXTRACTION

### 3.1. Introduction

As mentioned earlier, speech recognition from non-acoustic sensor data faces the problem of discovering an effective feature recognition strategy, and US + lip video SSIs, although attractive in many ways, share this drawback. Being based on images, their intrinsic input dimensionality may be of the order of 1 million pixels. Some means of dimension-reducing feature extraction is thus critical. (The following discussion is centered on tongue features. Lip features, which are much easier to handle, will for overall coherence be extracted in the same way as tongue features.)

### 3.2. Contour extraction approach

Tongue contour extraction is a tempting choice for reducing dimensionality that retains visual interpretability of the features. In ultrasound imaging of the tongue, the air-tissue boundary at the upper surface of the tongue produces a bright, continuous contour, referred to in a side-looking scan as the sagittal contour. Image processing tools for automatically extracting and characterizing this contour make ultrasound imaging a powerful tool for the study of speech production [53], [54]. Unfortunately, despite extensive literature on techniques for extracting tongue contours from ultrasound data (see [62]-[64] and references therein), tongue contour tracking remains an extremely challenging task. The high level of speckle noise in ultrasound images (multiplicative noise arising from the coherent nature of the ultrasound wave), coupled with variations in acoustic contact between the transducer and the speaker's skin; blocking of the ultrasound wave by the hyoid bone and jaw; poor reflectivity of muscle fibers in certain orientations of the tongue; and the lack of a complete echogenic tissue path to all parts of the tongue surface, in particular the tongue tip, often result in sagittal contours that are incomplete, contain significant artifacts, or are even totally absent. While even imperfect automatically-extracted contours remove the tedium of hand-scanning and are valuable for qualitative studies, it is difficult to integrate such information in a coherent way into labeled training datasets intended for machine learning tasks such as speech recognition. As a consequence, US-based SSIs have tended to use projective feature extraction techniques rather than contour finding. In work performed thus far, Principal Component Analysis, PCA, and the Discrete Cosine Transform, DCT, have been the methods of choice.

### 3.3. PCA and DCT approaches

In [8] and [10], PCA was used on ultrasound-based SSIs in an "Eigentongues" approach, wherein each ultrasound image is represented as a linear combination of a set of orthogonal Eigen-images determined on a training set of representative images. The first 30 or so Eigentongues were found sufficient to represent the discriminative power contained in the ultrasound images.

The DCT, widely used for lossy image compression, is based on the notion that most of the information in an image is concentrated in the lower spatial frequencies [65]. We note that the DCT, as a direct multiplicative transform related to the Fourier transform, does not make use of a training set. The technique for calculating the DCT coefficients will be presented in the next section. In [52], the article on which the SSC archive is based, it was found that DCT features provided substantially better recognition scores, as well as faster execution times, than the Eigentongue approach. **This result leads to the important consequence that the SSC benchmark refers to recognition scores obtained using DCT features** (we note in addition that the original Eigentongue features of [52] are no longer available). Consequently, a quantitative comparison of a DNN-HMM approach to the GMM-HMM analysis used in the original benchmark – which is the major impetus of this article – must make use of the identical DCT features in its baseline result.

The SSC archive DCT features were constructed in the following way. First, fixed Regions of Interest (ROI) of tongue and lip images were resized to 64 by 64 pixels. This resizing is necessary in order to keep the number of DCT coefficients tractable. For an image matrix $A$ of size $N*N$, the two-dimension DCT is then computed as:

$$D_{ij} = a_i a_j \sum_{m=0}^{N-1} \sum_{n=0}^{N-1} A_{mn} \cos\frac{\pi(2m+1)i}{2N} \cos\frac{\pi(2n+1)j}{2N} \quad (1)$$

$0 \ll i, j \leq N - 1$

where

$$a_i, a_j = \begin{cases} \frac{1}{\sqrt{N}}, & i, j = 0 \\ \sqrt{\frac{2}{N}}, & 1 \ll i, j \ll N - 1 \end{cases}$$

Dimensionality reduction is achieved by retaining only the *K* lowest frequency DCT components. In [52], a feature size of *K* = 30 was selected, based on performance comparisons. In acoustic speech recognition, it is usual to concatenate the first derivative, or Δ, of the MFCC feature vector to the vector itself. This procedure was also carried out for the DCT features of the archive, thus creating a 120-component feature vector for each tongue + lip frame.

### 3.4. New features created with Deep Auto Encoder

Although DCT features have provided promising recognition results for US + lip video SSIs, it has been necessary to make certain compromises in extracting them, notably: 1) resizing the original images before calculating them; and 2) retaining only a small, fixed number of DCT coefficients. While computational tractability issues prevent us, at present, from removing the first restriction, the presence of the raw tongue and lip data in the SSC archive allows us to consider taking a closer look at the second one.

It is first of all interesting to examine the appearance of tongue and lip images reconstructed using 30 DCT coefficients. An example result on 4 frames is given in figure 2.

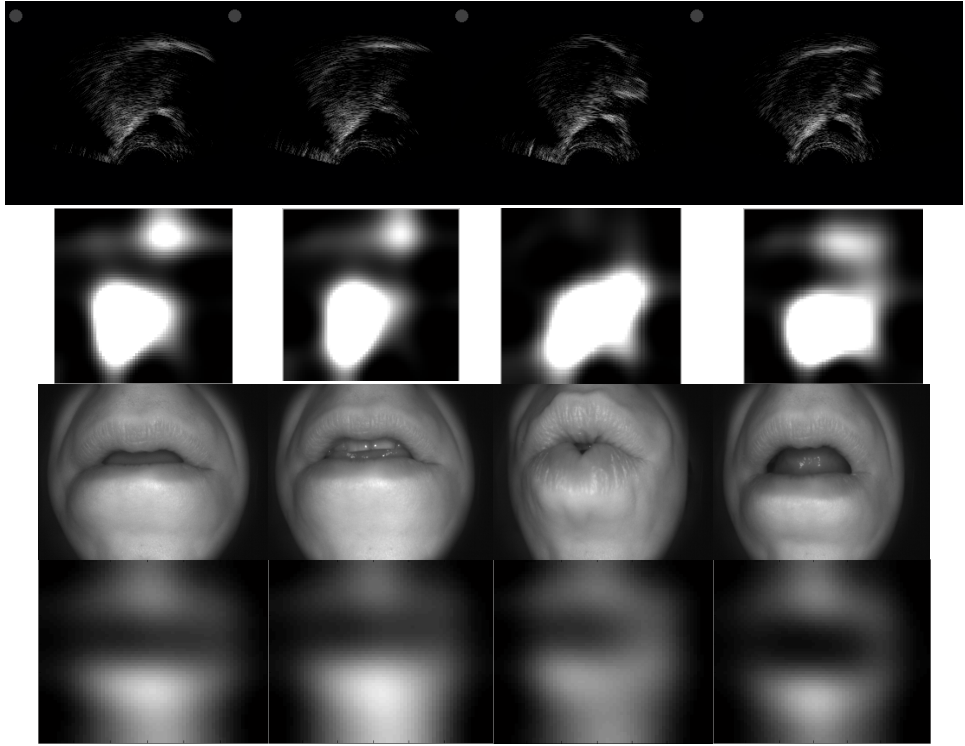

Figure 2: Original lip (top row) and tongue (third row) images compared to their reconstructions (second and fourth rows, respectively) using 30 DCT coefficients.

Although the lip reconstructions are sufficiently clear to distinguish an overall degree of mouth opening, an acoustically pertinent quantity, the visual fidelity of the tongue images is rather poor. The information in the tongue images necessary for distinguishing different acoustic configurations is, evidently, coded by the DCT in a way that does not retain a high level of visual fidelity. It is tantalizing to ask, however, whether one might do better by creating, from the original images present in archive, a new feature representation that reduces dimensionality while explicitly preserving visual fidelity, rather than relying on a somewhat arbitrarily placed cut in spatial frequency space, as was done for the DCT. A Deep Auto Encoder, or DAE, was used to explore this possibility.

A Deep Auto-Encoder is a neural network used for reducing the dimensionality and learning a representation of input data [66]. It contains an "encoder" and a "decoder" symmetrically arranged about a "code" layer, as shown in Figure 3. The action of the encoder can be defined as:

$$z = f(Wx + b) \qquad (2)$$

where $f$ is an activation function, such as sigmoid function, $W$ a weight matrix, and $b$ a bias. The decoder output is defined as:
$$x' = f(W'z + b') \qquad (3)$$
where $x'$ is of the same dimension as $x$. The weight matrix $W'$ is equal to $W^T$. An autoencoder is trained by minimizing the image reconstruction error, computed as:
$$L(x, x') = - \sum_{k=1}^{d}[f_1(x_k, x'_k) + f_2(x_k, x'_k)] \qquad (4)$$
where
$$f_1(x, x') = x \log x'$$
$$f_2(x, x') = (1 - x) \log(1 - x').$$
When training is complete, the code layer may be regarded as a compressed representation of the input, and is suitable for use as a feature vector. Details of the DAE training procedure can be found in [67].

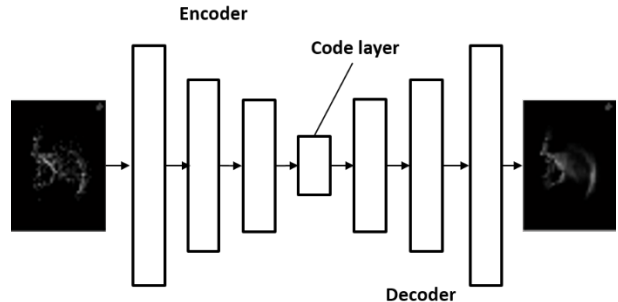

Figure 3. Architecture of DAE

To obtain the new features, ROIs were selected and resized, once again for computational tractability purposes, via bi-cubic interpolation, to 50×70 (lip) and 50×60 (tongue) pixel arrays, which form the inputs to the DAE. After tests with various architectures, a *J*-1000-500-250-*K* network was chosen, with *J* the number of inputs (3500 for lips and 3000 for the tongue); *K* the desired dimensionality of the created feature vector; and the intermediate figures the number of neurons in each layer. Features were calculated for *K* = 30, 20, 10, and 5. The encoder and symmetric decoder networks were trained on 12 lists of images selected at random from the SSC TIMIT training corpus.

Reconstructed images (bottom row of each panel) for tongue and lips are compared with the original images (top row of each panel) in Figure 4, where (a) and (b) show the results using 30 and 5 DAE features respectively. The figure shows that remarkable visual fidelity can be obtained using only 5 DAE features. This is in contrast to images reconstructed using DCT features shown previously, which are barely recognizable even for the 30 dimensional features. Although one may ask to what extent the DAE solution is similar to PCA, the $K = 5$ case, with, as will be seen later, the SSR results it allows to obtain, is nonetheless intriguing.

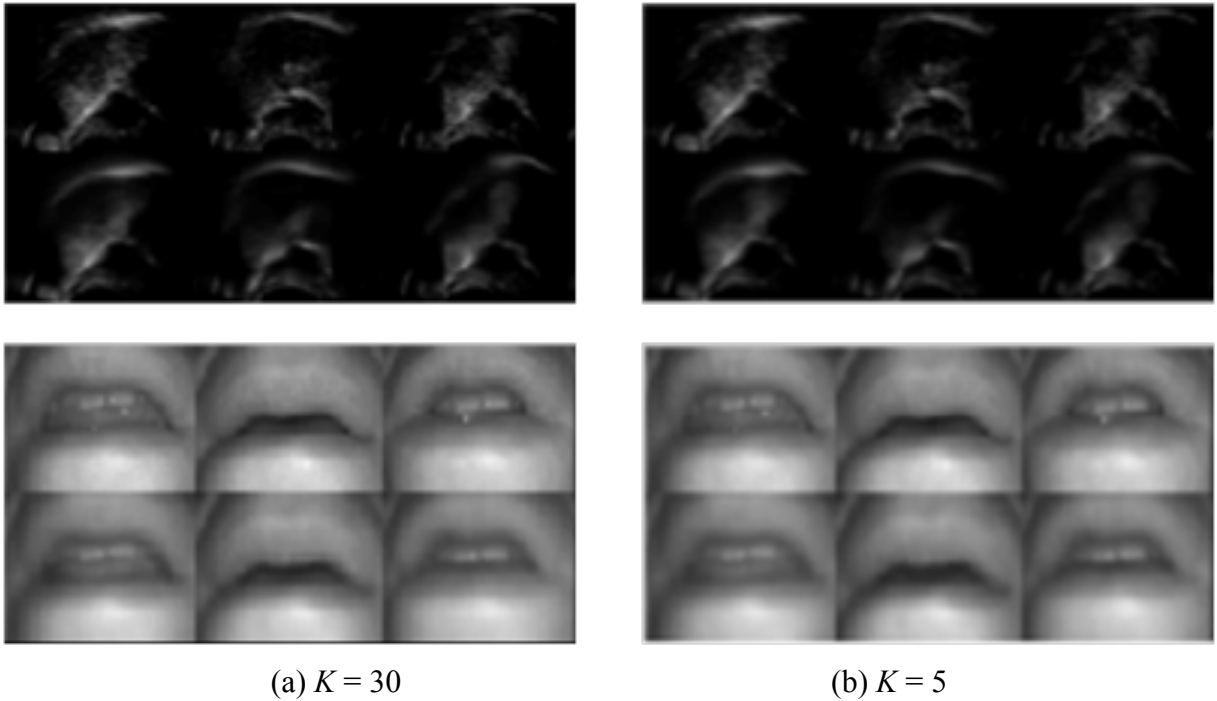

(a) $K = 30$  (b) $K = 5$

Figure 4: Original (top row of each panel) and reconstructed (bottom row of each panel) images of tongue and lips using two dimensionalities of DAE features.

**3.5. Feature-free approaches**

In the past few years, "feature-free" approaches to pattern recognition in speech, signal and image processing, based on convolutional neural networks, CNN, have proven very effective [68][69][70][71]. The CNN is a multilayer neural network consisting of multiple sub-networks with shared weights and overlapping receptive fields, alternated with "pooling" layers that reduce dimensionality by retaining only a subset of the afferent inputs. The use of shared weights across different instances of the identical sub-networks greatly reduces the number of weights to be learned, thus allowing the training of a CNN to remain relatively tractable. CNNs are thought to be able to learn a hierarchy of features, of progressively higher order as information pass from the input to the output of the network.

Recently, CNN have begun to make their entry into the field of SSI. In [72], which is actually a lip-reading application, a CNN is trained to transform video frames from a large video database directly into synthesized un-vocalized speech, using the video sound track to create source-filter type training labels. In [73], a CNN is trained to recognize phonetic targets in US tongue and video lip images in a 488-sentence single speaker database, using a phonetically labeled sound track as ground truth, for a speech recognition task with an HMM-GMM. In [74], CNN are used to recognize tongue gestural targets in US tongue images for an isolated phoneme + nonsense word recognition task. In the latter reference, extensive use is made of data augmentation [70] to increase the size of the training set, often a concern in using CNN, which require very large training sets to be effective, due to the large number of weights that must be learned.

Conceivably, the CNN technique could be applied to the raw images of the SSC archive, to try to improve on the ad-hoc DCT and DAE features tested thus far. As the archive contains no sound track, however, pre-training of the CNN, as in [72] and [73], will not be feasible: the CNN training will have to take place conjointly with that of the HMM probabilities. A study of this possibility will appear in an upcoming article.

## 4. DNN–HMM SPEECH RECOGNITION

### 4.1. System overview

The Kaldi open-source Deep Learning toolkit [75][76] was used to build the SSR system, whose overall architecture is illustrated in Figure 5. Features extracted from the archive data were first normalized to have zero mean and unit variance (Mean Variance Normalization (MVN) in the figure).
.

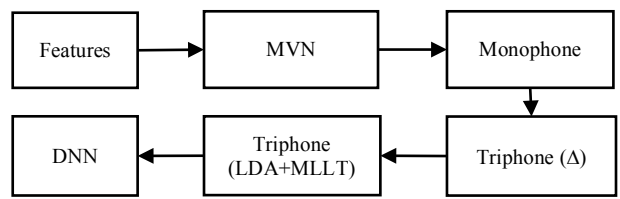

Figure 5: Overall SSR training procedure.

In the SSC benchmark, HTK was used to perform the speech recognition, using standard GMM-HMM architecture. In order to ensure as meaningful a comparison as possible with the benchmark result, without actually *re-doing* it with HTK, the recognition with Kaldi was performed first using a GMM-HMM, and then a DNN-HMM. The procedures used for the non-acoustic ASR, adapted from standard recipes in acoustic speech recognition and Deep Learning [77-81], are described below.

In the Kaldi GMM-HMM "acoustic" model training stage (the name "acoustic model" is retained even though the input feature data used here are non-acoustic), a monophone model was first trained using combined tongue and lip feature vectors, of type DCT or DAE, of dimension $K$ = 30, 20, 10, and 5. The alignment of the monophone is then used for the triphone1 training stage, where the Δ features are also included. In the subsequent phase, the triphone2b model is created using the alignment result of triphone1, and applying Linear Discriminant Analysis (LDA) and Maximum Likelihood Linear Transformation (MLLT) methods to replace the Δ features appearing in triphone1 and produce a new feature vector of dimension 40. The monophone, triphone1 and triphone2b acoustic models were trained consecutively, each time using the previous model for alignment, while for the DNN-HMM training, the alignment of triphone2b was used.

The Deep Belief Network (DBN) implemented is illustrated in Figure 6 [77-79], using the $D$ dimensional feature vectors as inputs. Restricted Boltzmann Machines (RBM) are cascaded by means of the weight vector $W$ in the figure. On the top layer of the DBN is a softmax output layer, and transition probabilities of the HMM are trained in the previous GMM-HMM phase. The system parameters are summarized in Table 1, including total numbers of Gaussians, tied state (regression tree), search space (beam), and acoustic/LM weight (acwt) parameters. The DNN training operates in two phases. During the pre-training phase, the RBMs are trained using the contrastive divergence (CD) algorithm. The 6 hidden layers of the RBMs are made up of Bernoulli-Bernoulli units (learning rate 0.4) except for the first, which is Gaussian-Bernoulli (learning rate 0.01). In the second phase, 90% of the training set was used to train the DNN, optimizing per-frame cross-entropy, and the remaining 10% of the training was used to test [80]. The weights learned in pre-training phase are then used to initialize the DNN model. The DNN architecture was implemented on a CUDA GPU machine.

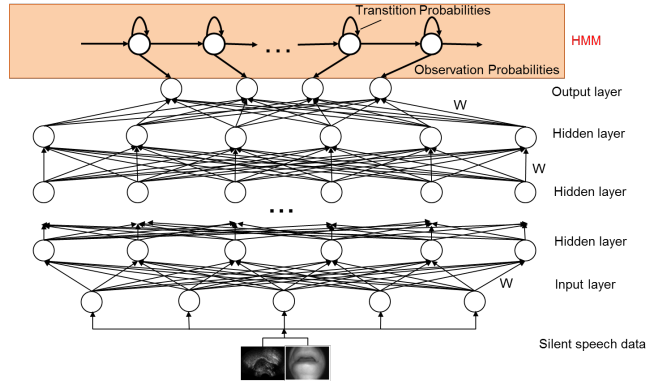

Figure 6. DNN structure used for SSR

Table 1. SSR system parameters

| Monophone | Tot_Gaussian | 1700 |
|---|---|---|
| Triphone1 | Regression tree leaves | 1800 |
| | Tot_Gaussian | 9000 |
| Triphone2b | Regression tree leaves | 3000 |
| | Tot_Gaussian | 25000 |
| DNN pretrain | Number of hidden layers | 6 |
| | Units per hidden layer | 1024 |
| DNN training | Number of hidden layers | 4 |
| | Units per hidden layer | 1024 |
| | beam | 13.0 |
| | Lattice_beam | 8.0 |
| | acwt | 0.1 |

**4.2. Language model and lexicon issues**

The lm_wsj_5k_nvp_2gram LM [61] used in the decoding stage of the SSC benchmark, derived from a fixed 5000-word subset of the Wall Street Journal (WSJ) text corpus, was also adopted in these tests. Obtaining realistic WER scores on small corpora, however, can be problematic. Using a closed vocabulary, as is the case here, tends towards underestimation of attainable WER. On the other hand, an unbiased lexicon derived exclusively from a small training set might not contain all the words present in the test set, thus leading to an overly pessimistic WER estimate. To help address these issues, a second estimate of the achievable WER on these data was also made using another, less task-specific LM, namely lm_csr_5k_nvp_2gram [61]. This LM contains newswire text from WSJ, the San José Meteor, and the Associated Press, along with some spontaneous dictation by journalists of hypothetical news articles. Results on both LM appear in the next section.

## 5. RESULTS AND ANALYSIS

Table 2 shows a comparison of the Kaldi DNN-HMM results, on the WSJ0 5000-word corpus, to those of the SSC benchmark, using the same 30-dimensional DCT input features and decoding strategy as in [52]. The formula used for WER is

$$WER = \frac{I+D+S}{N} \qquad (8)$$

where $I$ is the number of insertion error, $D$ is number of deletions, $S$ the number of substitutions, and $N$ the total number of words in the test set. Although a test with HTK itself was not repeated here, the fact that quite similar results were obtained using a GMM-HMM in Kaldi (column 2) provides reassurance that the figures obtained using Kaldi are reasonable. The Table shows that the DNN-HMM strategy has reduced the WER by almost a factor of 3 as compared to the benchmark.

Table 2. Comparison with original HTK result of [52], using 30-element DCT features

| Error | HTK SSC Benchmark | Kaldi GMM-HMM | **Kaldi DNN-HMM** |
|---|---|---|---|
| Insertion | 17 | 41 | **6** |
| Deletion | 23 | 17 | **8** |
| Substitution | 138 | 120 | **52** |
| Number of words | 1023 | 1023 | **1023** |
| Correct words | 862 | 886 | **963** |
| Correct rate | 84.26% | 86.61% | **94.13%** |
| WER | 17.4% | 17.4% | **6.45%** |

To perform the LM tests proposed in section 4.2, the procedure was repeated using the alternate LM, as shown in Table 3 for 30-element feature vectors of both types (DCT and DAE). One notes first of all that the DCT and DAE features give similar performances, barring the monophone case. We will return to this point in the discussion of Tables 4 and 5. Nonetheless, although the WER performance obtained on the less specific LM is somewhat worse, as expected, it is still significantly better than the SSC benchmark, for both types of features.

Table 3. Comparing results for the 2 different LM, for 30-element feature vectors of both types

| LM | | WER (%) | |
|---|---|---|---|
| | | lm_csr_5k_nvp_2gram | lm_wsj_5_nvp_2gram |
| DCT | monophone | 45.55 | 40.47 |
| | Triphone2b | 17.40 | 12.71 |
| | **DNN** | 11.44 | **6.45** |
| DAE | monophone | 58.55 | 59.92 |
| | Triphone2b | 21.70 | 14.76 |
| | **DNN** | 13.98 | **7.72** |

To further explore different types of input features, DCT and DAE feature vectors of dimension $K = 20$, 10, 5 for each visual modality were also tested. Results are given in Table 4 for the WSJ LM, and in Table 5 for the CSR LM. Overall, higher scores are again obtained with the more task specific LM, as expected. The tables also show that for both LM, similar results are obtained for the two types of features, with the DCT being slightly better, when the dimensionality $K$ of the input vectors is 10 or more. For $K = 5$, however, while the DCT features are no longer salient, the DAE retains most of its effectiveness. Thus, although the DAE has not been completely successful at simultaneously optimizing saliency and low dimensionality, the results it furnishes are intriguing, and suggest that it may be possible to do better with a more sophisticated approach.

Table 4. Recognition results with WSJ LM

| WER of | DCT (%) | | | |
|---|---|---|---|---|
| Dimension | 30 | 20 | 10 | 5 |
| Monophone | 40.47 | 37.15 | 36.36 | 98.24 |
| Triphone2b | 13.00 | 14.76 | 12.32 | 100 |
| **DNN** | **6.45** | **6.35** | **7.43** | **99.51** |
| WER of | DAE (%) | | | |
| Dimension | 30 | 20 | 10 | 5 |
| Monophone | 59.92 | 44.18 | 41.15 | 45.45 |
| Triphone2b | 14.76 | 14.96 | 15.54 | 17.79 |
| **DNN** | **7.72** | **7.72** | **8.80** | **10.07** |

Table 5. Recognition results with CSR LM

| WER of    | DCT (%) |       |       |       |
|-----------|---------|-------|-------|-------|
| Dimension | 30      | 20    | 10    | 5     |
| Monophone | 45.55   | 40.86 | 39.78 | 98.34 |
| Triphone2b| 17.79   | 19.16 | 16.42 | 100   |
| **DNN**   | **11.44**| **11.53** | **12.32** | 99.80 |
| WER of    | DAE (%) |       |       |       |
| Dimension | 30      | 20    | 10    | 5     |
| Monophone | 58.55   | 52.00 | 49.76 | 54.25 |
| Triphone2b| 21.70   | 21.41 | 19.75 | 22.48 |
| **DNN**   | **13.98**| **13.10** | **14.37** | **14.86** |

## 6. CONCLUSIONS AND PERSPECTIVES

A confrontation of the SSC recognition benchmark with DNN-HMM SSR techniques using the Kaldi Deep Learning package has led to an improvement in WER of almost a factor of 3 in the most favorable scenario, thus helping to establish US as a highly attractive SSI modality. Tests performed using both the original WSJ LM and a less task-specific CSR LM give WER values that are on these data, using the significantly improved compared to the benchmark. Before the DNN-HMM tests, Kaldi was also used to test a GMM-HMM architecture, in order to demonstrate compatibility with the methods used in the benchmark. New features derived from the raw benchmark data using a DAE give results only slightly worse than those obtained with the original DCT features, while retaining their effectiveness even at very low dimensionality. Both new and original features have now been appended to the SSC benchmark data.

While these results are promising, SSR still remains somewhat less accurate than acoustic speech recognition, and further work will be necessary. In the future, for the SSC benchmark, it will be interesting to experiment with other feature extraction strategies, for example convolutional neural networks, CNN, which might allow the image-resizing step, where information may be lost, to be skipped. For SSI more generally, it will be interesting to accumulate much larger (if possible multi-speaker) data sets, so that some of the mentioned problems associated with small speech data sets may be avoided.

## 7. ACKNOWLEDGEMENTS

This research was supported by the National Nature Science Foundation of China (No. 61303109 and No. 61503277) and 985 Foundation from China's Ministry of Education (No. 060-0903071001).